\documentclass[final,12pt]{colt2022} 

\title[Noise-Free Kernel-Based Bandit]{Open Problem: Regret Bounds for Noise-Free Kernel-Based Bandits}
\usepackage{times}

\medskip
\coltauthor{%
 \Name{Sattar Vakili} \Email{sattar.vakili@mtkresearch.com}\\
 \addr MediaTek Research
}

\def\Oc{\mathcal{O}}
\def\Oct{\tilde{\mathcal{O}}}

\usepackage{setspace}

\usepackage[utf8]{inputenc} 
\usepackage[T1]{fontenc}    
\usepackage{hyperref}  
\hypersetup{
    colorlinks=true,
    linkcolor=blue,
    citecolor =blue,
    filecolor=magenta,      
    urlcolor=magenta,
}
\usepackage{url}            
\usepackage{booktabs}       
\usepackage{amsfonts}       
\usepackage{nicefrac}       
\usepackage{microtype}      
\usepackage{lipsum}
\usepackage{graphicx}

\usepackage{amsmath}
\usepackage{amsfonts}
\usepackage{algorithm}
\usepackage{algorithmic}
\usepackage{bm}
\usepackage{amssymb}
\usepackage{color}
\usepackage{xcolor}
\usepackage{hyperref}
\usepackage{amsmath, amssymb, graphicx, url}

\usepackage{cleveref}
\crefformat{section}{\S#2#1#3} 
\crefformat{subsection}{\S#2#1#3}
\crefformat{subsubsection}{\S#2#1#3}

\usepackage{mathrsfs}
\usepackage{makecell}

\def\Ib{\bm{I}}

\def\Hc{\mathcal{H}}

\def\TP{\top}

\def\argmax{\text{argmax}}

\def\Kb{\mathbf{K}}
\def\kb{\mathbf{k}}

\def\yb{\mathbf{y}}
\def\Ib{\mathbf{I}}

\def\Rr{\mathbb{R}}

\def\E{\mathbb{E}}

\def\Xc{\mathcal{X}}
\def\Rc{\mathcal{R}}
\def\Oc{\mathcal{O}}
\def\Oct{\tilde{\mathcal{O}}}

\newtheorem{assumption}{Assumption}

\begin{document}

\maketitle

\begin{abstract}%
  Kernel-based bandit is an extensively studied black-box optimization problem, in which the objective function is assumed to live in a known reproducing kernel Hilbert space. While nearly optimal regret bounds (up to logarithmic factors) are established in the noisy setting, surprisingly, less is known about the noise-free setting (when the exact values of the underlying function is accessible without observation noise). We discuss several upper bounds on regret; none of which seem order optimal, and provide a conjecture on the order optimal regret bound. 

\end{abstract}

\begin{keywords}%
Kernel-based bandit, Bayesian optimization, Gaussian processes
\end{keywords}

\section{Introduction}

Black-box optimization of a (possibly non-convex) function from expensive evaluations is a ubiquitous problem in machine learning, including both academic research and industrial applications. Those include A/B testing, hyperparameter tuning \citep[including AlphaGo,][]{chen2018bayesian}, robotics, environmental monitoring, and more \citep{Shahriari2016outofloop}. A particularly successful approach is based on the use of Gaussian process modeling, due to its versatility, modeling power, and ability to provide uncertainty estimates. A mathematical formulation of the problem under the bandit setting has been extensively studied in the literature. Nearly optimal regret bounds are established in the noisy setting. Surprisingly, the problem of order optimal regret bounds remains open under the noise-free setting. Here, we overview the existing results and give a formal description of the open problem, aspiring to motivate solutions.

\section{Problem Setup}\label{sec:setup}

Consider the sequential optimization of an objective function $f:\Xc\rightarrow\Rr$, where $\Xc\subset\Rr^d$ is a compact set. A learning algorithm is allowed to collect a sequence of observations $\{(x_i, y_i)\}_{i=1}^\infty$, where $y_i=f(x_i)$ in the noise-free setting, and $y_i=f(x_i)+\epsilon_i$ in the noisy setting with $\epsilon_i$ being a well behaved observation noise. The objective is to get as close as possible to the maximum of~$f$. The performance of the algorithm is measured in terms of (cumulative) regret, defined as the cumulative loss in the values of the objective function at observation points, compared to a global maximum:
\begin{eqnarray}
\Rc(N) = \sum_{i=1}^N \left(f(x^*) - f(x_i)\right), 
\end{eqnarray}
where $x^*\in \argmax_{x\in \Xc} f(x)$ is a global maximum.  The regularity assumption on $f$ which makes the problem tractable is given next. 

\paragraph{The RKHS and the regularity assumption on $f$:} Consider a positive definite kernel $k:\Xc \times \Xc\rightarrow \Rr$ with respect to a finite Borel measure. Let $\Hc_k$ denote the reproducing kernel Hilbert space (RKHS) corresponding to $k$, defined as a Hilbert space equipped with an inner product $\langle.,.\rangle_{\Hc_k}$ satisfying the following: $k(.,x)\in \Hc_k$, $\forall x\in \Xc$, and $\langle f,k(.,x)\rangle_{\Hc_k}=f(x)$, $\forall x\in\Xc, \forall f \in \Hc_k$ (reproducing property). The typical assumption in kernel-based models is that the objective function~$f$ satisfies $f\in\Hc_k$ for a known kernel $k$. 
\begin{assumption}\label{ass_norm}
Assume $f$ is fixed with $\|f\|_{\Hc_k}\le C_k$, for some $C_k>0$, where $\|f\|_{\Hc_k}=\sqrt{\langle f,f\rangle_{\Hc_k}}$. 
\end{assumption}


Under Assumption~\ref{ass_norm}, the sequential optimization problem given above is often referred to as that of kernel-based (kernelized) bandit, Gaussian process (GP) bandit, or Bayesian optimization. The latter two terms are motivated by the algorithm design which often employs a GP surrogate model. 

\paragraph{GP surrogate  model:}

It is useful for algorithm design to employ a zero-mean GP model $F$ with kernel $k$, which provides a surrogate posterior mean (prediction) and a surrogate posterior variance (uncertainty estimate) for the kernel-based model.  Defining $\mu_n(x) = \E\big[F(x)|\{(x_i,y_i)\}_{i=1}^{n}\big]$ and $\sigma_n^2(x) = \E\big[(F(x) - \mu_n(x))^2|\{(x_i,y_i)\}_{i=1}^{n}\big]$, it is well known that
$
\mu_n(x) =  \kb_n^{\TP}(x)  (\lambda^2 \Ib_n+\Kb_n)^{-1}\yb_n$ and 
$\sigma_n^2(x) = k(x,x) - \kb_n^{\TP}(x)  (\lambda^2 \Ib_n+\Kb_n)^{-1}\kb_n(x), 
$
where
$\yb_n=[y_1,\dots, y_n]^{\top}$,
$\kb_n(x) = [k(x,x_1), \dots, k(x,x_n)]^{\TP}$, $\Kb_n$ is the positive definite kernel matrix $[\Kb_n]_{i,j} = k(x_i, x_j)$
, $\Ib_n$ is the identity matrix of dimension $n$, and $\lambda^2$ is the variance of a zero mean Gaussian surrogate distribution for the noise. In the noise-free setting, one may set $\lambda=0$ in the above expressions. 

The Mat\'ern family of kernels is perhaps the most commonly used in practice~\citep[see, e.g.,][]{snoek2012practical, Shahriari2016outofloop} and theoretically interesting~\citep[see, e.g.,][]{ srinivas2010gaussian,bull2011convergence} family of kernels. It is known that the RKHS corresponding to a Mat\'ern kernel is equivalent to a Sobolev space~\citep[see, e.g.,][]{Teckentrup2019} that provides an intuitive interpretation of the function class based on its smoothness. Furthermore, the Mat{\'e}rn kernels may be insightful for the theory of neural networks, as it has been shown that the neural tangent kernel~\citep{jacot2018neural} is equivalent to a Mat{\'e}rn kernel~\citep{vakili2021uniform}. Following the literature, we also emphasize the Mat{\'e}rn family in the formulation of the open problem.

\section{Overview of the Results in the Noisy Setting}


We first overview the main results for the kernel-based bandit problem in the noisy setting and under Assumption~\ref{ass_norm}. 
Classical algorithms such as GP upper confidence bound \citep[GP-UCB,][]{srinivas2010gaussian, Chowdhury2017bandit}, Thompson sampling~\citep[GP-TS,][]{Chowdhury2017bandit}, and expected improvement~\citep[EI,][]{gupta2022regret} sequentially select the observation points based on a score referred to as {acquisition function}. The best known regret bounds for these algorithms
scales as $\Oct(\Gamma_{k,\lambda}(N)\sqrt{N})$ over $N$ steps (see the references above), where $\Gamma_{k,\lambda}(N)$ is a kernel specific complexity term, 
referred to as the \emph{maximal information gain} between the noisy observations and the latent GP surrogate model~\citep{srinivas2010gaussian}:
\begin{eqnarray}
\Gamma_{k,\lambda}(n) =\sup_{\{x_i\}_{i=1}^n\subset\Xc}\frac{1}{2}\log\det(\Ib_n+\frac{1}{\lambda^2}\Kb_n).
\end{eqnarray}
That is also nearly equivalent to the \emph{effective dimension} of the kernel for a dataset of finite size $n$~\citep{Calandriello2019Adaptive}. Kernel specific bounds on $\Gamma_{k,\lambda}(n)$, with a focus on Mat{\'e}rn family, are provided in~\cite{srinivas2010gaussian, vakili2020information, vakili2021uniform, kassraie2022neural}. 

The $\Oct(\Gamma_{k,\lambda}(N)\sqrt{N})$\footnote{The notations $\Oc$ and $\Oct$ are used for mathematical order, and that up to hiding logarithmic factors, respectively.} scaling is not tight in general, and may even fail to be sublinear in many cases of interest, since $\Gamma_{k,\lambda}(N)$ may grow faster than $\sqrt{N}$. It remains an open problem whether the suboptimal regret bounds of these acquisition based algorithms is a fundamental limitation or a shortcoming of their proof~\citep[see][for the details]{vakili2021open}.

On discrete domains, the SupKernelUCB algorithm was shown to have an $\Oct(\sqrt{\Gamma_{k,\lambda}(N)N})$ regret bound~\citep{Valko2013kernelbandit}.
SupKernelUCB is not considered to be practical. Several more practical algorithms (which also apply to continuous domains) with $\Oct(\sqrt{\Gamma_{k,\lambda}(N)N})$ regret have been proposed recently: a tree-based domain-shrinking algorithm~\citep[GP-ThreDS,][]{salgia2021domain}, Robust Inverse Propensity Score for experimental design~\citep[RIPS,][]{camilleri2021high}, batched pure exploration~\cite[BPE,][]{li2021gaussian}, and its sparse version~\citep[S-BPE,][]{vakili2022improved}.

Substituting the bounds on $\Gamma_{k,\lambda}(N)$ for the Mat\'ern kernels leads to $\Rc(N)=\Oct(N^{\frac{\nu+d}{2\nu+d}})$, where $\nu$ is the smoothness parameter of the kernel. \cite{Scarlett2017Lower} proved a matching (up to logarithmic factors) $\Omega(N^{\frac{\nu+d}{2\nu+d}})$ lower bound on the regret, which shows the order optimality of the $\Oct(\sqrt{\Gamma_{k,\lambda}(N)N})$ regret bounds in this case. 


\section{Regret Bounds in the Noise-Free Setting}

As outlined above, there are many results on the kernel-based bandit problem under the noisy setting, several of which achieving nearly optimal regret bounds. Under the noise-free setting,
\cite{lyu2019efficient} is the only work providing bounds on the cumulative regret, while \cite{bull2011convergence} provided nearly optimal bounds on the simple regret. In this section, we first overview these two works, and then provide a formal description of the open problem of cumulative regret under the noise-free setting. 

\subsection{From Noisy to Noise-Free}

One straightforward way to establish regret bounds under the noise-free setting is to set the noise equal to zero in the results under the noisy setting. 
\cite{lyu2019efficient} took this approach and showed an
$\Oc(\sqrt{\Gamma_{k,\lambda}(N) N})$ bound on the cumulative regret for GP-UCB. 
The analysis mainly follows that of GP-UCB under the noisy setting~\citep{srinivas2010gaussian}, except for using tighter confidence intervals for GP models under the noise-free setting. In the case of Mat{\'e}rn kernels, this implies an $\Rc(N)=\Oct(N^{\frac{\nu+d}{2\nu+d}})$ regret that always grows faster than $\sqrt{N}$, even for highly smooth kernels (large $\nu$). As we will see next, this regret bound is not order optimal in the noise-free setting.

\subsection{Explore then Commit}

Consider the simple regret variation of the kernel-based bandit problem. In this variation, after $n$ (possibly adaptive) observations $\{(x_i,y_i)\}_{i=1}^{n}$, the algorithm selects a candidate maximizer $\hat{x}_n$, and its performance is measured in terms of simple regret defined as $r_n=f(x^*)-f(\hat{x}_n)$. 
Focusing on Mat\'ern family and simple regret, \cite[][]{bull2011convergence} proved an $r_n=\Oct(n^{-(\min\{\nu,1\})/d})$ regret bound for the EI algorithm (see their Theorem $2$), that translates to an $R(N)=\Oc(N^{(d-1)/d})$ bound on its cumulative regret when $\nu\ge 1$ (which is typically the case, except for the Laplace kernel where $\nu=\frac{1}{2}$).
Furthermore, they showed that EI mixed with pure exploration achieves
an $r_n=\Oct(n^{-\frac{\nu}{d}})$ simple regret (see their Theorem $5$), which cannot be improved (see their Theorem~$1$). 
The cumulative regret $R(N)$ for this algorithm, however, grows linearly in $N$, as a consequence of pure exploration. The question of cumulative regret appears more challenging due to the
exploration-exploitation tradeoff, which is inherent to the bandit problems.

Inspired by the results of \cite{bull2011convergence}, and using the \emph{explore then commit} technique~\citep[e.g., see][]{garivier2016explore}, we can design a simple algorithm with sublinear regret. Specifically, consider an algorithm which performs pure exploration by, e.g., choosing nearly uniform points across the domain, up to step $N_0$. The algorithm then commits to the best observation point based on the GP prediction: $\hat{x}_{N_0}=\arg\max_{x\in\Xc}\mu_{N_0}(x)$ (with $\lambda=0$ in the expression of $\mu_{N_0}$), for the remaining of the steps: $x_i=\hat{x}_{N_0}, \forall i>N_0$ . Choosing the optimum value for $N_0$, leads to a cumulative regret of $\Rc(N)= \Oc(N^{\frac{d}{\nu+d}})$. Note that this regret bound is tighter than the one in~\cite{lyu2019efficient} given above, indicating that the analytical techniques used in the noisy setting may not be suitable for the noise-free setting.


\subsection{Open Problem}

\paragraph{Problem~1}
Consider the kernel-based bandit problem given in Section~\ref{sec:setup} under the noise-free setting ($\epsilon_i=0, \forall i$). What is the lowest growth rate of $R(N)$ with $N$? 
As a specific case of interest, when the kernel $k$ is a Mat{\'e}rn kernel with smoothness parameter $\nu$ and $R(N)=\Oct(N^{\alpha})$, what is the smallest value of $\alpha$, achievable by a learning algorithm?

\subsection{Discussion}

In the case of Mat{\'e}rn kernels, as outlined above, the existing upper bounds include $\Oct(N^{\frac{\nu+d}{2\nu+d}})$~\citep{lyu2019efficient}, $\Oc(N^{\frac{d-1}{d}})$~\citep[EI,][]{bull2011convergence}, and $\Oct(N^{\frac{d}{\nu+d}})$ (explore then commit). We however conjecture that the following regret bound is achievable:
\begin{eqnarray}
\Rc(N)= 
\begin{cases}
    \Oc(N^{(d-\nu)/d}),~~~ &\text{when}~d>\nu,\\
    \Oc\left(\log (N)\right), &\text{when}~d=\nu,\\
    \Oc(1), &\text{when}~d<\nu.
\end{cases}
\end{eqnarray}


We here give an informal reasoning for this claim. Following standard UCB-based bandit algorithm techniques, it can be shown that the GP-UCB
algorithm attains $\Rc(N)=\Oc(C_k\sum_{i=1}^N\sigma_{i-1}(x_i))$. Let $\{x^*_i\}_{i=1}^n\in\arg\max_{\{x_i\}_{i=1}^n\subset \Xc}\sum_{i=1}^n\sigma_{i-1}(x_i)$ and $\Theta^*_n= \sum_{i=1}^n\sigma_{i-1}(x^*_i)$. We then have $\Rc(N)=\Oc(C_k\Theta^*_N)$, which reduces the problem to bounding $\Theta^*_n$. We conjecture that under certain mild regularity of the domain, $\{x^*_i\}_{i=1}^n$ are distributed nearly uniformly across the domain, in which case $\sigma_{i-1}(x^*_i)=\Oc(i^{-\frac{\nu}{d}})$~\cite[see, e.g.,][Lemma~$3.8$]{Teckentrup2019}. That, when summed over $i$, leads to $\Theta^*_n=\Oc(n^{\frac{d-\nu}{d}})$ when $d>\nu$, $\Theta^*_n=\Oc\left(\log(n)\right)$ when $d=\nu$, and $\Theta^*_n=\Oc(1)$ when $d<\nu$.



\bibliography{references}

\appendix



\end{document}